\documentclass{article}

\usepackage[preprint]{corl_2024} 

\usepackage{subcaption}
\usepackage{amssymb}
\usepackage{amsmath}
\usepackage{graphicx}
\usepackage{wrapfig}
\usepackage{import}
\usepackage{pgfplots}
\usepackage{tikz}
\usepackage{enumitem}
\usepackage{booktabs}
\usepackage{lipsum}
\usepackage{appendix}
\usepackage{spverbatim}

\title{DegustaBot: Zero-Shot Visual Preference Estimation for Personalized Multi-Object Rearrangement}

\usepackage{xcolor}

\author{
  Benjamin A.~Newman$^{1*}$~~Pranay Gupta$^1$~~Kris M.~Kitani$^1$~~Yonatan Bisk$^1$\\ 
  \textbf{Henny Admoni$^1$~~Chris Paxton$^2$}\\
  $^1$ Carnegie Mellon University~~$^2$ Hello Robot\\
  $^*$\textit{Corresponding Author: Benjamin Newman,~newmanba@cmu.edu}
}

\begin{document}
\maketitle


\begin{abstract}
\textit{De gustibus non est disputandum} (``there is no accounting for others' tastes'') is a common Latin maxim describing how many solutions in life are determined by people's personal preferences. Many household tasks, in particular, can only be considered fully successful when they account for personal preferences such as the visual aesthetic of the scene.
For example, setting a table \emph{could} be optimized by arranging utensils according to traditional rules of Western table setting decorum, without considering the color, shape, or material of each object, but this may not be a completely satisfying solution for a given person. Toward this end, we present DegustaBot, an algorithm for visual preference learning that solves household multi-object rearrangement tasks according to personal preference. To do this, we use internet-scale pre-trained vision-and-language foundation models (VLMs) with novel zero-shot visual prompting techniques. To evaluate our method, we collect a large dataset of naturalistic personal preferences in a simulated table-setting task, and conduct a user study in order to develop two novel metrics for determining success based on personal preference. This is a challenging problem and we find that 50\% of our model's predictions are likely to be found acceptable by at least 20\% of people.

\end{abstract}



\section{Introduction}
\label{sec:introduction}

Assistive robots in homes should complete tasks in ways that align with user preferences~\citep{newman2022helping}. 
Consider the task of setting out dishes and utensils on a table for dinner at your home. The \textit{correct} solution for this task is subject to your preferences for the type of utensils you want to use and how you prefer to arrange them. However, modeling individual preferences, especially those that consider fine-grained features such as the real-valued location of objects and their relative placements, is challenging as such preferences are subjective, difficult to specify explicitly, and vary from person to person. In this work, we explore this problem through the example of personalized dinner table arrangement, and determine a personalized task plan for setting a dinner table given examples of a person's preference.


Prior research on multi-object rearrangement collects problem-specific datasets of simulated or human demonstrations that represent personal preferences for completing a household task. They hypothesize that collecting  large, task-specific preference datasets is sufficient to train a feature space that can generalize to novel participants who were excluded from the training data~\citep{abdo2015robot, kang2018automated, kapelyukh2022my, newman2023towards}. These methods, however, often yield poor generalization to these held-out participants due to large effects from individual differences. These challenges indicate that is it difficult to collect a large dedicated dataset to train a feature space that covers the unbounded space of personal preferences for even a single task and suggests that generalized pretraining is a potentially fruitful path of research. These ideas are further compounded when considering underspecified preferences and preferences that are influenced by cultural norms or decorum.


\begin{figure}[t]
  \centering
  \includegraphics[width=1\linewidth]{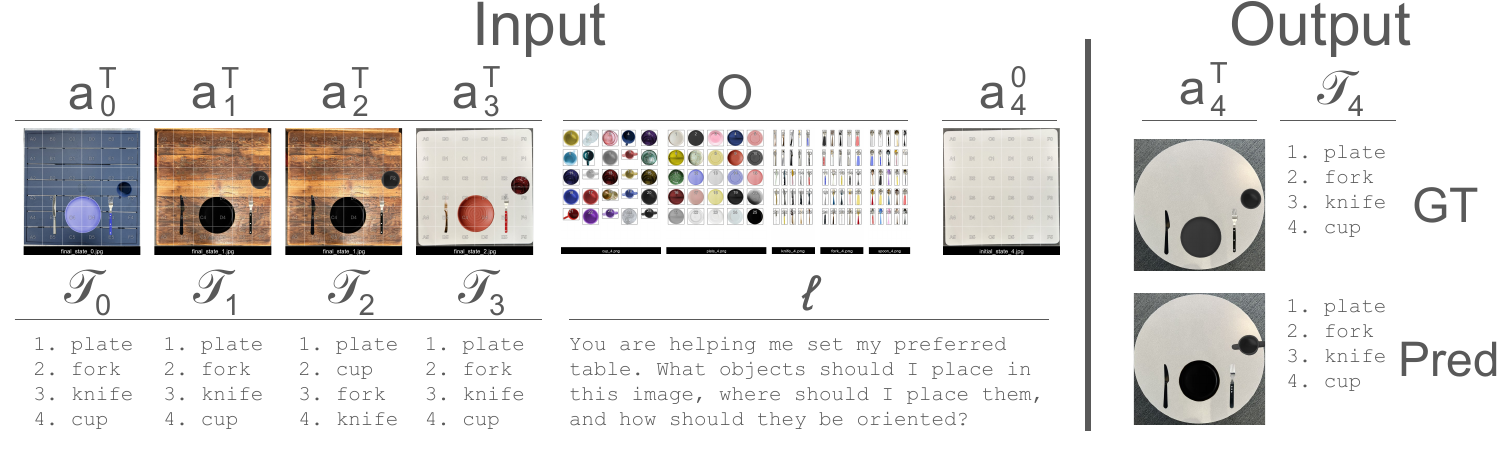}
  \caption{DegustaBot takes in a single person's preferred table arrangements (shown here as the visual context $k_i$ and order of object placement $\mathcal{T}_i$), the objects $O$ from which the algorithm can select, the table to set $a_0$ and the task prompt $\ell$. The robot then produces a task plan in the form of an object arrangement (visualized as an image $a_T$) and the order of object placement $\mathcal{T}$. This predicted arrangement, Pred, should match a held-out preference created by the user, GT.}
  \label{fig:system_diagram}
\end{figure}

Large-language models (LLMs) may provide a solution. They are pretrained on internet scale data, and have shown the ability to solve tasks upon which they weren't explicitly trained.
They can leverage knowledge about the abstract external concepts that often guide personal preferences. In fact, it has been shown they can infer generalizable user preferences from few examples with in-context learning~\citep{wu2023tidybot}. Representing multi-object preference learning solely in language, however, removes important visual information about a task. At the same time, these problems cannot be solved \textit{purely} visually, as they do not have a grounded vocabulary of actions over which to reason.
Therefore, we developed a new prompting technique that provides such a vocabulary. We built a virtual tabletop setting task, and collected a large dataset of human preferences for different arrangements in order to test this technique. Concretely, in this paper, we make the following contributions:
\begin{itemize}
[noitemsep,leftmargin=15pt]
    \item DegustaBot: a novel method for implicit visual preference learning to find personalized solutions to fine-grained multi-object rearrangement tasks. 
    \item A novel human evaluation used to evaluate personalized multi-object rearrangement agents.
    \item A dataset of naturalistic preference data in a simulated multi-object rearrangement task. 
\end{itemize}
\section{Related Work}
\label{sec:related_work}
 \noindent\textbf{Multi Object Rearrangement:}
Recent advances in embodies AI have led to a flurry of benchmarks where an embodied agent is tasked with rearranging objects in a real or simulated home environment~\citep{puig2018virtualhome, batra2020rearrangement, newman2020optimal, szot2021habitat, ehsani2021manipulathor, weihs2021visual, gan2021threedworld, newman2022harmonic}. However, these benchmarks have specified goal states, as compared to our work where the preferred object arrangement has to be inferred from context examples. 

Another line of work, focused on object rearrangement, where-in the target objects are specified via pointing gestures~\citep{rasch2019tidy}, eye gaze~\citep{newman2020examining}, or target layouts~\citep{yan2021quantifiable} during task execution. However these works require manually specified target location for every object, and are hence human effort intensive. A line of follow-up works address this issue by modeling preferences using learnt priors about where objects typically go~\citep{taniguchi2021autonomous, kant2022housekeep, sarch2022tidee}. These preferences however are generic are not personalized. 

Prior works on personalized object rearrangement, rely on simulated or large crowd sourced datasets of human preferences to learn fixed latent preference vectors~\citep{kapelyukh2022my} or latent preferences that can be adapted online~\citep{newman2024bootstrapping}, model spatial relationships~\citep{kang2018automated} or perform collaborative filtering~\citep{abdo2015robot}. In contrast, our approach leverages in-context learning with large-scale pretrained VLMs to perform personalized object rearrangement. While ours is not the first work which leverages foundational models to perform few shot personalized rearrangement, we are first ones to perform fine-grained preference alignment, i.e, spatial preferences. As compared to prior works~\citep{wu2023tidybot}, which operate over a discrete preference space, i.e, identifying the correct receptacle for each object.

\noindent\textbf{Foundation Models for Robotics:} VLMs pre-trained on large scale datasets have shown commonsense reasoning abilities. Researchers have leveraged these abilities to perform planning and control for robotics~\citep{firoozi2023foundation, hu2023toward}. Many prior works~\citep{song2023llm, huang2022language, saycan2022arxiv, huang2022inner, zeng2022socratic, liu2023reflect, raman2022planning, lin2023text2motion, liu2023llm+, wang2023describe, silver2022pddl, newman2024leveraging} have used pre-trained LLMs to generate actionable natural language plans for robots. VLMs have also been used to generate subgoals for navigation~\citep{dorbala2022clip, chen2023open, shah2023lm, shah2023navigation, gadre2023cows, huang2023visual} and manipulation~\citep{cui2022can, shridhar2022cliport} tasks. Additionally, prior works have also leveraged LLMs to directly generate low-level executable policy code for robots~\citep{liang2023code, singh2023progprompt}. Another line of works, has also used LLMs to generate rewards, which can be for RL~\citep{huang2023voxposer, yu2023language, ma2023eureka}. In our work, we use a VLM to generate the policy code to accomplish a continuous preference aligned novel goal state.

\noindent \textbf{Visual Prompting in Vision Language Models}
The development of in-context learning for few shot adaption of LLMs~\citep{brown2020language}, was followed up by a flurry of prompt optimization approaches. While one line of work focuses on prompt-tuning~\citep{lester2021power} or prefix-tuning~\citep{li2021prefix} through backpropagation using numerical gradients, others rely on the generated answer scores to perform optimization with textual gradients ~\citep{pryzant2023automatic}, prompt search using genetic algorithms~\citep{xu2022gps}, or iterative prompt refinement~\citep{yang2023large}. More recently, approaches have tried to use visual prompting to improve VLM performance by leveraging a VLM's ability to solve multiple choice problems.~\citep{nasiriany2024pivot, liu2024moka}. Our approach adapts these methods to a complex visual preference learning task using these insights to prompt VLMs to combine information from multiple images to reason about a person's preference. 


\section{Problem Specification}
\label{sec:problem_domain}


People have widely varying preferences, making it difficult to develop one-size fits all solutions, even for problems that at first glance seem to be well-specified. One person may consider a table to be arranged with a simple fork and a bowl, while another won't sit down to eat until their flower arrangement is perfect and a candle is lit. Furthermore, these preferences cannot easily be specified in natural language: language is often either ambiguous
or requires cumbersome description.

\begin{wrapfigure}[16]{R}{0.5\textwidth}
    \vspace{-10pt}
  \centering
  \includegraphics[width=1\linewidth]{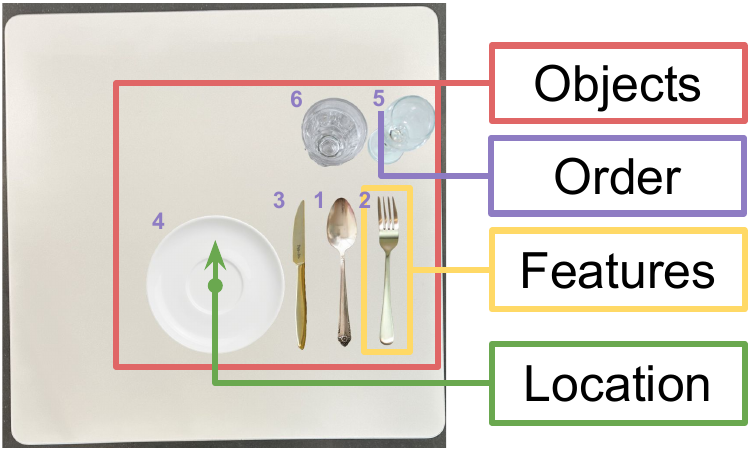}
  \caption{Details of a table arrangement. An arrangement is described by the objects within the arrangement, the order in which they are placed, their features (such as color, shape and material, and their location and orientation. }
  \label{fig:system_diagram}
\end{wrapfigure}

%
%
Instead, we argue that preferences can be specified implicitly by visual representations of the final state of a multi-object rearrangement task. 


To investigate this, we choose table setting as an example application. 
Concretely, the goal of a preference model $\mathcal{M}$ is to maximize the likelihood of seeing a table arrangement $a \in A$ given an initial state $a_0$, a representation of a preference $K = \left[k_0, k_1, \dots, k_C\right]$ each element of which is a concatenation of the arrangement and object placement order $k_c = \left[a_c^T, \mathcal{T}_c\right]$, a set of objects that can be used for the arrangement $O$ and a language prompt $\ell$ that describes the general task: $\max_{a\in A} \mathcal{M}(a~|~K, O, a_0, \ell)$. 

An arrangement $a$ of length $T$ is a sequence of tuples containing positioning $\mathbf{x}$ and object $\mathbf{o}$, that is, $a_{1:T} = [(\mathbf{x}_1, \mathbf{o}_1), \cdots, (\mathbf{x}_T, \mathbf{o}_T)]$. $\mathbf{x}~\in~\mathbb{R}^3$ represents the continuous elements of an arrangement (x and y-positions of an object on the table's surface and r, the object's rotation), and $\mathbf{o}$ represents the discrete elements of an arrangement (an object's color $c$, material $m$, and type $u$). 
When represented in language, an object can be described by a multi-label vector of up to two colors (beige, black, blue, brown, gold, gray, green, pink, purple, red, silver, white, and yellow), two materials (ceramic, glass, metal, plastic, wood), and a single type (cup, fork, knife, plate, or spoon). 

As in prior work,~\citep{nasiriany2024pivot}, we consider a family of functions $\Omega$ that lift non-visual information into the image domain. An arrangement can be lifted into the visual domain with $\Omega_a: a \rightarrow I_a \times \mathcal{T}_a$, where $I_a$ is a visual representation of the arrangement and $\mathcal{T}$ is a language description of the order in which objects were placed. We also consider $\Omega_O: O \rightarrow I_O$, which jointly maps all objects into the visual domain. 

Finally, we consider a preference to be implicitly defined by a history of previous arrangements completed by a single person. In a slight abuse of notation, we assume that $\Omega_a(K) = [\Omega_a(a_0), \Omega_a(a_1), \cdots, \Omega_a(a_C)]$.


\textbf{Objective.}
In this work, we want to investigate whether foundation models trained on internet scale data and conditioned on implicitly defined preferences can produce personalized task plans. In other words, we consider the following problem:
\[\mathcal{M}_\theta(K, O, a_0, \ell) \stackrel{?}{=} \mathcal{M}_{H}(O, a_o, \ell)~\text{where}~K~\sim \mathcal{M}_H,\]

where $\mathcal{M_\theta}$ is a model parameterized by parameters $\theta$, and $\mathcal{M}_{H_k}$ is a human with preference $k$. 
We parameterize $M_\theta$ as a vision and language model. While there have been a surfeit of large foundation models recently becoming available, we choose to investigate three high performing, recent models: OpenAI's GPT-4o~\citep{openai2023gpt4}, Anthropic's Claude-3 Haiku~\citep{anthropicteam2024anthropic}, and Google's Gemini 1.5 Pro~\citep{geminiteam2024gemini}. 


\section{Estimating Simulated Table Arrangements using VLMs}
\label{sec:method}

Solving our implicit, visually represented table-top arrangement task requires incorporating visual information across several images (e.g. $k$, $o$, $a_0$).
We present and evaluate several different zero-shot methods for prompting VLMs to produce task plans to that solve table-top arrangement problems. We do this through \textit{lifting functions} which add extra information to an image before sending it to a VLM like GPT4v alongside a query, such as adding a grid or highlighting a point in space.


\subsection{Producing Task Plans through Visual Prompting}

We evaluate different lifting functions $\Omega_a$ and $\Omega_O$ for solving zero-shot, personalized table-top arrangement. In addition to lifted images, we include a text prompt that describes the problem  and the desired format of the output. 
Prompting specifics are left to the appendix.

\begin{figure}[t]
  \centering
  \includegraphics[width=1\linewidth]{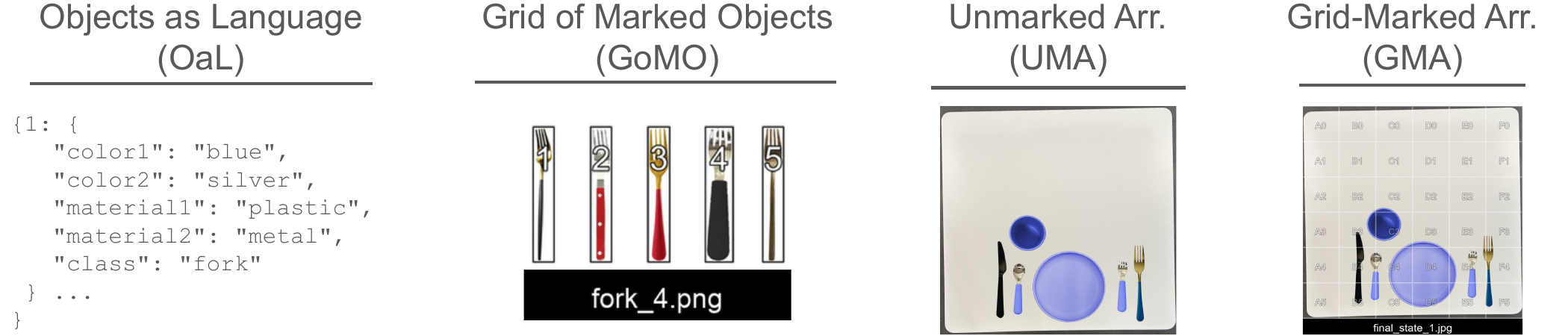}
  \caption{Object and arrangement lifting functions, from left to right: OaL, a representation of objects as language, for this we use the json representation; GoMO, grid of marked objects, which represents objects visually with referential marks overlaid on each object; UmA, Unmarked Arrangement lifts the arrangement into the image domain; and finally GMA, Grid-Marked Arrangement, which overlays a spatial reference grid on the continuous table top space. }
  \label{fig:lifting}
\end{figure}

\subsubsection{Selecting Objects}
We first describe the object lifting functions, $\Omega_O$:

\textbf{Objects as Language} We take each object in our dataset and represent it as a dictionary of features, which are concatenated to create a dictionary of all objects and their representations (see Fig.~\ref{fig:lifting}, left).

\textbf{Grid-of-Marked Objects.} We represent objects visually by creating grids of objects within a category. Each object is overlaid with a spatial reference marker to aid in object detection~\citep{yang2023setofmark} and the image is labeled with object type and preference number (see Fig.~\ref{fig:lifting}, second column).


\subsubsection{Selecting Locations and Rotations}
We represent arrangements visually through  $\Omega_a$ in two ways: 

\textbf{Unmarked Arrangements.} We assume that positions $\textbf{x}$ represent an object's centroid. Then, representative images are resized to fit within an initial arrangement (e.g. an empty table) and are pasted at $\textbf{x}$. They are then rotated according to the rotation $\textbf{r}$. 

\textbf{Grid-Marked Arrangements.} We turn predicting a continuous location into a multiple choice question over grid cells that serve as spatial reference markers. First, we overlay a spatial grid on each preference image and then ask the model to produce all the grid cells that an object will intersect with on a table after it is placed. We average the centroids of the responses over multiple responses to get a continuous position prediction.



\subsubsection{Combining Lifting Functions into Visual Prompting Methods}

We combine these lifting functions into four methods that test each models ability to predict task plans in table-top arrangement tasks as follows: 
\begin{itemize}
[noitemsep,leftmargin=15pt]
    \item \textbf{DegustaBot-LOUMA.} uses Language Objects with UnMarked Arrangements
    \item \textbf{DegustaBot-LOGMA.} uses Language Objects with Grid-Marked Arrangements
    \item \textbf{DegustaBot-MOUMA.} uses Marked Objects with UnMarked Arrangements
    \item \textbf{DegustaBot-MOGMA.} uses Object Images with Grid-Marked Arrangements.
\end{itemize}

\subsection{Simulating Preferences for Table Arrangement}

We develop a small dataset of simulated preferences for controlled evaluation. These preferences are drawn from three positional preferences, three object preferences, and a single order preference. We design these preferences to approximate a standard, a semi-standard and a non-standard table setting. To test our model over a distribution of positions, we add positional and rotational noise to these table settings. The object preferences are also drawn from three color preferences: a preference for all red, all blue, or all yellow objects.



We displayed arrangements on two visually distinct tables. In total, we simulated 18 preferences on two table tops for a total of 36 preferences. We tested these in both a reconstruction task (predict on the same table) and a generalization task (predict on a different table) for 72 total experiments. 

\subsection{Validation Metrics}

We test two main hypotheses:
(1) that large vision and language models develop a feature space that is useful for estimating preferences for real-valued multi-object rearrangement tasks; and 
(2) that visually grounding preference information aids in estimating complex preferences.
%
%

\textbf{Object selection:} this measures a method's ability to select objects consistent with the preference. We compute this as the accuracy of object selection: did the model select exactly the right object? 

\textbf{Position Selection:} We take a geometric approach to comparing arrangements. While a point-to-point comparison seems immediately intuitive, the geometric relationship between points in a particular table setting are important, as well. For example, a table settings performing poorly when measured with point-to-point correspondences may still be similar geometrically to the preferred arrangement, only displaced slightly. To account for these issues, we introduce another evaluation metric: root mean squared deviation (RMSD).


We use the Kabsch algorithm~\citep{Kabsch:a12999} to register matched points between predicted and ground truth arrangements, and find the translation vector $g$ and rotation matrix $r$ that minimize the root mean squared deviation between the two paired frames:

\[\min_{r \in R, g\in G} \sqrt{\frac{1}{N}\sum_n^N (\bar{A}^{gt}_n - R(\bar{A}^{pred}_n - g))^2 }.\]

We report this minimized RMSD as our geometric evaluation metric. 


\subsection{Results}
\label{sec:results}

\begin{figure*}[t]
    \begin{subfigure}[t]{.49\linewidth}
    \centering
    \includegraphics[width=1\linewidth]{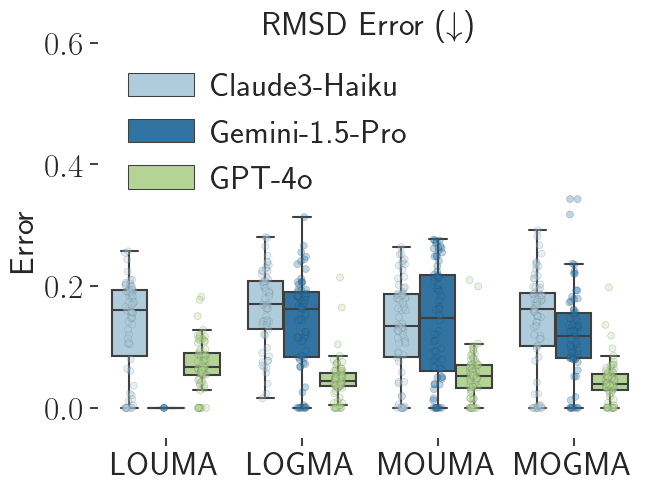}
    \end{subfigure}
    \hfill
    \begin{subfigure}[t]{.49\linewidth}
    \centering
    \includegraphics[width=1\linewidth]{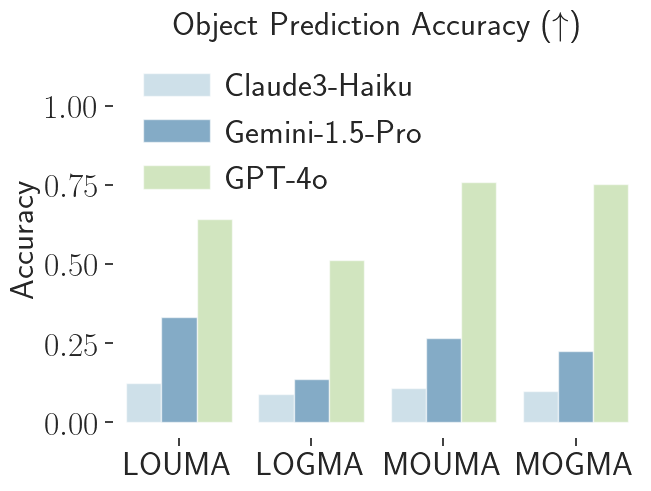}
    \end{subfigure}
    \hfill
    \caption{Quantitative results for evaluating DegustaBot on simulated preferences. Left shows the performance of each model's ability to capture the geometry of the table arrangement, as measured by RMSD. Right shows each model and method's accuracy when choosing items to place in the arrangement. GPT-4o and MOGMA perform the best on both metrics.}
    \label{fig:sim_results}
\end{figure*}

We test each model with each combination of visual lifting functions and report results on how well the predicted image matches the ground truth scene's geometry, as reported through RMSD, and chosen utensils, as reported through accuracy. These results are shown in Fig. \ref{fig:sim_results}. 

First, we see that GPT-4o outperforms other choices of VLM in both RMSD and object prediction, under all methods. We can also see that grid-marked methods tend to improve a model's ability to match scene geometry. Finally, we see that representing utensils visually aids in correctly predicting utensils. Results for Gemini-1.5-Pro, LOUMA are missing due to a 97\% failure rate on this method.   
Taking these results together, we see that DegustaBot-MOGMA using GPT-4o outperforms all other methods on this task. We will use this method to analyze naturalistic preferences in the next section. 




\section{Naturalistic Tabletop Arrangement Dataset}


To understand if this approach works on naturalistic data, we collected a dataset of preferences for table arrangements through an online study. 
We then assessed model performance, as in Sec.~\ref{sec:results}.


\subsection{Data Collection}

Our user study consisted of a task with two stages: \textit{Preference Elicitation} and \textit{Self-Evaluation}. During Preference Elicitation, we prompt participants to create their preferred table-top arrangements. In Self-Evaluation, participants rate these arrangements, as well as several transformations of their  arrangements. We used these ratings to develop a subjective analysis of model performance. Participants saw a total of six rounds of this task: a practice round, followed by five experimental trials. 

We collected a dataset of 995 table arrangements drawn from 199 participants. Participants were paid USD\$4.00 for completing the 20 minute long task. All participants were located within the United States at the time of task completion. We did not collect further demographic information about our participants. This study was approved by the IRB of record.

\subsubsection{Preference Elicitation}

Participants were first presented with an image of one of five different empty tables, and 
asked to to set the table by selecting objects from five object categories: plates, spoons, forks, knives, and cups. After selecting an object category, participants were shown all instances in that category simultaneously. Each category contained 25 objects of varying shapes, colors, and materials, yielding an object dataset of 125 creative commons licensed and public domain images of table setting utensils. 

\begin{wraptable}{tr}{.6\textwidth}
    \vspace{-10pt}
    \begin{tabular}{rccc}
    \toprule  
    RMSD Threshold                    &  0.01           &   0.05     &  0.10            \\ \cmidrule(lr){1-4}
    \% of difference ratings $< 0.2$  &  0.727          &   0.333    &  0.199           \\
    \% of model responses             &  0.145          &   0.332    &  0.532           \\ \bottomrule
    \end{tabular}
    \caption{We compare our method's performance against subjective acceptance at several RMSD thresholds. When RMSD is below 0.01, 72.7\% of people are likely to find a scene acceptable. 14.5\% of the scenes our model produced fell into this category. At 0.05 RMSD, 33.3\% of people find scenes acceptable, while 33.2\% of scenes our model produced met this threshold. Finally at 0.10 RMSD, 19.9\% of scenes were found to be acceptable, while our model produce such scenes 53.2\% of the time.}
    \label{tab:subject_acceptance}
\end{wraptable}

Participants were then presented with a simple drag and drop interface through which they could place the object anywhere on the table's surface and rotate it object to their liking. Participants continued this procedure until they were satisfied with the arrangement. The only restriction placed on table arrangements was that participants were required to set a minimum of three objects. After completing a table arrangement, participants moved on to the self-evaluation portion of the task. 

\subsubsection{Subjective Acceptability}

We then asked participants to evaluate several arrangements.

\textbf{Rating the initial scene.} After creating an arrangement, $a_{intial}$ participants were prompted to answer the question, \textit{Please rank your agreement to the following statement: This is my preferred table arrangement}, sliding scale from 0-100 to offer their baseline acceptability score $b_0$.

\textbf{Jittering the initial scene.} To understand what errors affect people's sense of an acceptable arrangement we randomly transformed $a_{intial}$. A transformation magnitude between 0 and 1 was randomly selected as $t_j$. The scene $a_{intial}$ was then translated and rotated by this amount to create $a_{jitter}$. Participants then gave a second acceptability rating $b_{jitter}$. 



\begin{figure}[t]
  \centering
  \includegraphics[width=1\linewidth]{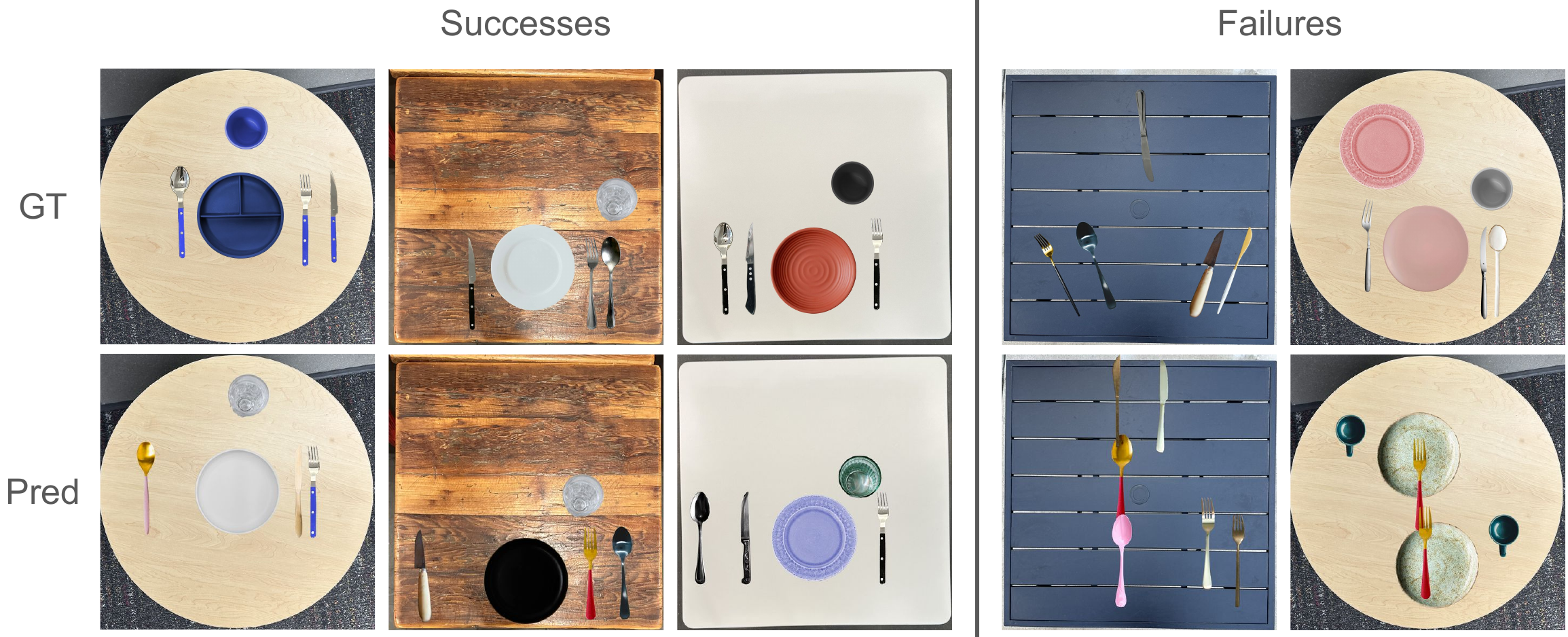}
  \caption{Qualitative results. On the top line we see ground truth images from our naturalistic preference data and on the bottom we see DegustaBot's predictions, lifted into the image domain. On the left side of the image we see three examples where DegustaBot predicts similar arrangements to the ground truth image. On the right we see some failure cases.}
  \label{fig:qualitative}
\end{figure}

\textbf{Threshold of Acceptability.}
We take the difference in ratings $b_{initial} - b_{jitter}$, and correlate this with the distance between $a_{initial}$ and $a_{jitter}$, giving us a \textbf{subjective acceptability} measure. We choose a ratings difference threshold of 0.2, which is interpreted as the threshold at which people find $a_{jitter}$ to be at least as acceptable as $a_{intitial}$. We then report the percentage of people who meet this acceptability threshold at various levels of RMSD. This tells us the percentage of people who are likely to find a scene acceptable below a given RMSD threshold. 

\subsection{Results}

We evaluated our method on naturalistic preference data. In Tab.~\ref{tab:subject_acceptance} we report the rate at which our model produces acceptable tables as measured by the RMSD distance between the scene our model produces and a held out arrangement. From these data we can see that 1) people are highly sensitive the the jittered scenes we produced, making this a strict metric and 2) our model produces scenes that people would likely find acceptable under this metric between 14.5\% and 53.2\% of the time. 

In Tab.~\ref{tab:acc_and_context} we report object prediction accuracy and RMSD as the number of preferences images increases from 0, 2, to 4. We see that increased context length improves performance metrics, but potentially with diminishing returns. This indicates that 1) human preferences are diverse and not easily predicted in a context-free setting and 2) our method makes efficient use of few context examples to predict a person's preference. We also see that the performance here is well below the performance on the synthetic data, indicating that visual preference learning from naturalistic preference data is a difficult task.

\begin{wraptable}{tr}{.6\textwidth}
    \vspace{-10pt}
    \begin{tabular}{rccc}
    \toprule  
    Length of k                    &  0           &   2      &  4             \\ \cmidrule(lr){1-4}
    Object Prediction Accuracy ($\uparrow$)  &  0.037          &   0.090    &  0.089           \\
    RMSD ($\downarrow$)             &  0.174          &   0.116    &  0.108           \\ \bottomrule
    \end{tabular}
    \caption{We report object prediction accuracy and RMSD of our method on naturalistic preferences. We see that with increased context length our method achieves higher performance, though perhaps with diminishing returns. }
    \label{tab:acc_and_context}
\end{wraptable}

Finally in Fig.~\ref{fig:qualitative} we show several qualitative results. Each column depicts a different participant within our dataset. The top row displays ground truth images, and the bottom rows depict our method's prediction. When geometric patterns are regular, our method does well at predicting them.






\section{Limitations}
In this work, we developed a method that can serve as a preference perception pipeline for a robot that performs household multi-object rearrangement tasks, DegustaBot. While we do not perform this experiment on a real robot, this setting emulates suction cup pick-and-place routines that have been shown to work reliably~\citep{kapelyukh2024dalle}. We instead focus on generating the task plan for such a task from realistic preference data. 


Additionally, we assumed that a robot has access to a person's prior table setting arrangements. 
Part of our contribution is to release a large dataset of realistic table-top arrangements for others to train from. In reality, the amount of data that we require from an individual to achieve reasonable performance is quite small compared to prior preference learning approaches~\citep{kapelyukh2022my}. 

A final limitation is that VLMs are not trained for fine-grained rearrangement tasks like ours. We presented several prompting methods that achieve reasonable performance even under this limitation, but our method would improve substantially if the underlying models were fine-tuned for fine-grained image reasoning tasks or even trained from the ground up for this kind of problem~\citep{chen2024spatialvlm}. 


\section{Conclusion}
\label{sec:conclusion}

We introduced DegustaBot, a novel method for prompting VLMs to perform the challenging task of visual preference learning in multi-object rearrangement tasks. We formalize this task, and then analyze several methods' performance over simulated preferences. We then collect a large dataset of naturalistic preferences and evaluate our best performing method, finding that although this is a challenging task, our method produces acceptable table arrangements.


\clearpage
\acknowledgments{If a paper is accepted, the final camera-ready version will (and probably should) include acknowledgments. All acknowledgments go at the end of the paper, including thanks to reviewers who gave useful comments, to colleagues who contributed to the ideas, and to funding agencies and corporate sponsors that provided financial support.}


\bibliography{main}  

\newpage
\appendix
\appendixpage
\section{Further Details on Prompting and Predicting}
Here we present further details on how we prompt VLMs to obtain task plans for complex, fine-grained preference tasks. Specifically, complex, fine-grained preference tasks include those in which a preference can be held over multiple different features of the task simultaneously and these features can be discrete, such as the color or material of an object, or continuous, such as the precise location of an object on a table's surface. 

\subsection{Selecting Objects}

\textbf{Objects as Language.} Our first lifting function represents objects as language. We do this by taking each object in our dataset, and coding them for color, material, and type of utensil. We take these features and create an easy to parse json file. Then, we give each object an arbitrary ID between 0 and 125 and concatenate the ID and language representation into a json file with 125 elements, each of which is a different object representation. We include this json as a part of the prompt to the VLM. To make a prediction about which object to choose, the VLM predicts an object ID directly. When we request multiple responses from the model and choose the mode of the model responses as the predicted object.

\begin{verbatim}
example_json = {
    "color1": "blue", "material1": "ceramic",
    "shape": "round", "pattern": "solid",
    "texture": "smooth", "class": "cup",
    "material2": "ceramic", "color2": "blue",
}
\end{verbatim}

\textbf{Grid-of-Marked Objects.} We turn object selection into a multiple-choice question answering task. We do this by displaying all the objects that the model could use to arrange the table simultaneously in images separated by object category. Each object instance is overlaid with a spatial reference mark. The model uses these marks to identify which object to place. Again, since we ask for multiple responses, we take the mode of the responses as our method's prediction. 

\subsection{Predicting Locations and Rotations}
\textbf{Unmarked Arrangements.} To produce an arrangement using this lifting function, we prompt the model to produce directly an x-position, y-position and rotation in the normalized image coordinate frame, e.g. (0,1). The final answer is an average over model responses.

\textbf{Grid-Marked Arrangements.} Our second arrangement lifting function again borrows the idea of inserting spatial reference markers into a scene for both localization and vocabulary grounding. We turn this task into a visual multiple choice question-answering task by overlaying a spatially marked grid using grid cells labelled with numbers and letters. To make a prediction,  We then take the weighted average of the locations of the centers of the grid cells and use this as our prediction of $\mathbf{x}$. To predict a rotation $r$, we discretize the rotation space into eight cardinal directions and ask the model to produce the one that is closest to the direction the object should be pointing, assuming the top of the image is North. Again, we average over all model responses to produce a final answer. 

\subsection{Predicting Placement Order}
We don't consider any visual lifting functions to predict object order. We simply include the placement order for each arrangement in $k$ in the problem definition text prompt passed to the VLM. In this problem definition, we also ask the model to produce its response list ordered by the placement order of each object. To find the ultimate task plan, we first take the mode of the lengths of each response to get the task plan length. Then, for each step in the task plan, we take the mode over the utensil type. 

\subsection{Prompts}

We use the following prompts for each of our model variants:

\textbf{Objects as Language}
\begin{spverbatim}
My preferences for setting a table are shown in the first images. You are helping me set the table in the final image according to my preferences. What objects should I place in this image, where should I place them, and how should they be oriented? Give a position in [x,y] where each value is a number between 0 and 1. Give rotation as the number of degrees clockwise from image north. Give your answer as a python list formatted as follows: [{'type':object_type, 'id': object_reference_id, 'position': [x, y], ‘rotation’: degrees}]. Give object_id as an integer. Include only this list in your response.
\end{spverbatim}

\textbf{Unmarked Arrangements}
\begin{spverbatim}
My preferences for setting a table are shown in the images named final_state_K.jpg. You are helping me set the table in initial_state_1.jpg according to my preferences. What objects should I place in this image, where should I place them, and how should they be oriented? Give a position in [x,y] where each value is a number between 0 and 1. Give rotation as the number of degrees clockwise from image north. Give your answer as a python list formatted as follows: [{`type':object_type, `id': object_reference_id, `position': [x, y], `rotation': degrees}]. Give object_id as an integer. Include only this list in your response.
\end{spverbatim}

\textbf{Grid Marked Arrangements} \begin{spverbatim}
My preferences for setting a table are shown in the images named final_state_N.jpg. You are helping me set the table in initial_state_1.jpg according to my preferences. What objects should I place in this image, where should I place them, and how should they be oriented? Please list all grid cells the object will intersect after it is placed. Give the orientation as the closest cardinal direction [N, NE, E, SE, S, SW, W, NW] the object is pointing in after it is placed. The top of the image is N. Give your answer as a python list formatted as follows: [{`type':object_type, `id': object_reference_id, `position':[grid_cell_ids], `cardinal_direction':direction}]. Give grid_cell_ids as a list of string. Give object_id as an integer. Include only this list in your response. 

\end{spverbatim}

\section{Further Details on the Kabsch Algorithm}

To compute the RMSD between two arragnements, we first define each arrangement as a frame of points: $A^{gt}$ and $A^{pred}$. We match points between the predicted and ground truth frames by computing their similarity matrix $D$, using the Hamming distance over features as our distance function. Then for each object in the ground truth frame, we find the most similar object in the predicted setting, ensuring that each object only matches with one object from the other setting. This yields two orderings $\bar{A}^{gt}$ and $\bar{A}^{pred}$ over the points in the respective frames. We ensure that each ordering is of the same length, $N$.

\section{VLM Model Choices and Costs}

We factor in several considerations when choosing the VLMs to evaluate. First, we want to optimize for the performance of the model; we want to test the best available models to test our hypothesis that they can effectively featurize complex preferences. However, technical limitations do play a role in our eventual decisions. For example, the token request limit and pricing for larger Anthropic models (Sonnet and Opus) made them prohibitive to use for this work.  

\begin{table*}[ht]
    \centering
    \begin{tabular}{rcccccc}
    \toprule 
     & \multicolumn{2}{c}{Input} & \multicolumn{2}{c}{Output} &  \multicolumn{2}{c}{Total} \\ \cmidrule(lr){2-3}  \cmidrule(lr){4-5}  \cmidrule(lr){6-7}
                    &  Tokens   &  USD\$ & Tokens   &  USD\$ & Tokens & USD\$   \\ \cmidrule(lr){1-7}
    GPT-4o   &  8035	            &  0.05      &   6250                &  0.10     &     14285    &     0.15 \\
    Claude Haiku      &  40175	            &  0.02      &   6250                &  0.01      &     46425     &     0.03   \\
    Gemini 1.5 Pro Vision     &   $-$         &  0.08 &       $-$        &  0.02 &  $-$ & 0.10    \\\bottomrule
    \end{tabular}
    \caption{We compare the costs of running various models for a single experiment in our dataset. A single experiment consists of 4 arrangement images, 4 text labels for the placement order, and a text prompt. We ask each model for 5 total responses. This is handled automatically by OpenAIs API. For Anthropic and Google models, we manually pass 5 requests. Google's pricing model prices per image and character, rather than tokens, so this column is omitted. Costs are rounded up to the nearest whole cent.}
    \label{tab:costs}
\end{table*}

In Tab.~\ref{tab:costs} we show the various token and dollar costs to run our method on a single experiment in our dataset. While we ran these experiments prior to Google's implementation of a pricing model, we include estimated costs here, too.

\section{Examples of the Simulated Task}

\begin{figure*}[t]
    \centering
    \includegraphics[width=0.75\linewidth]{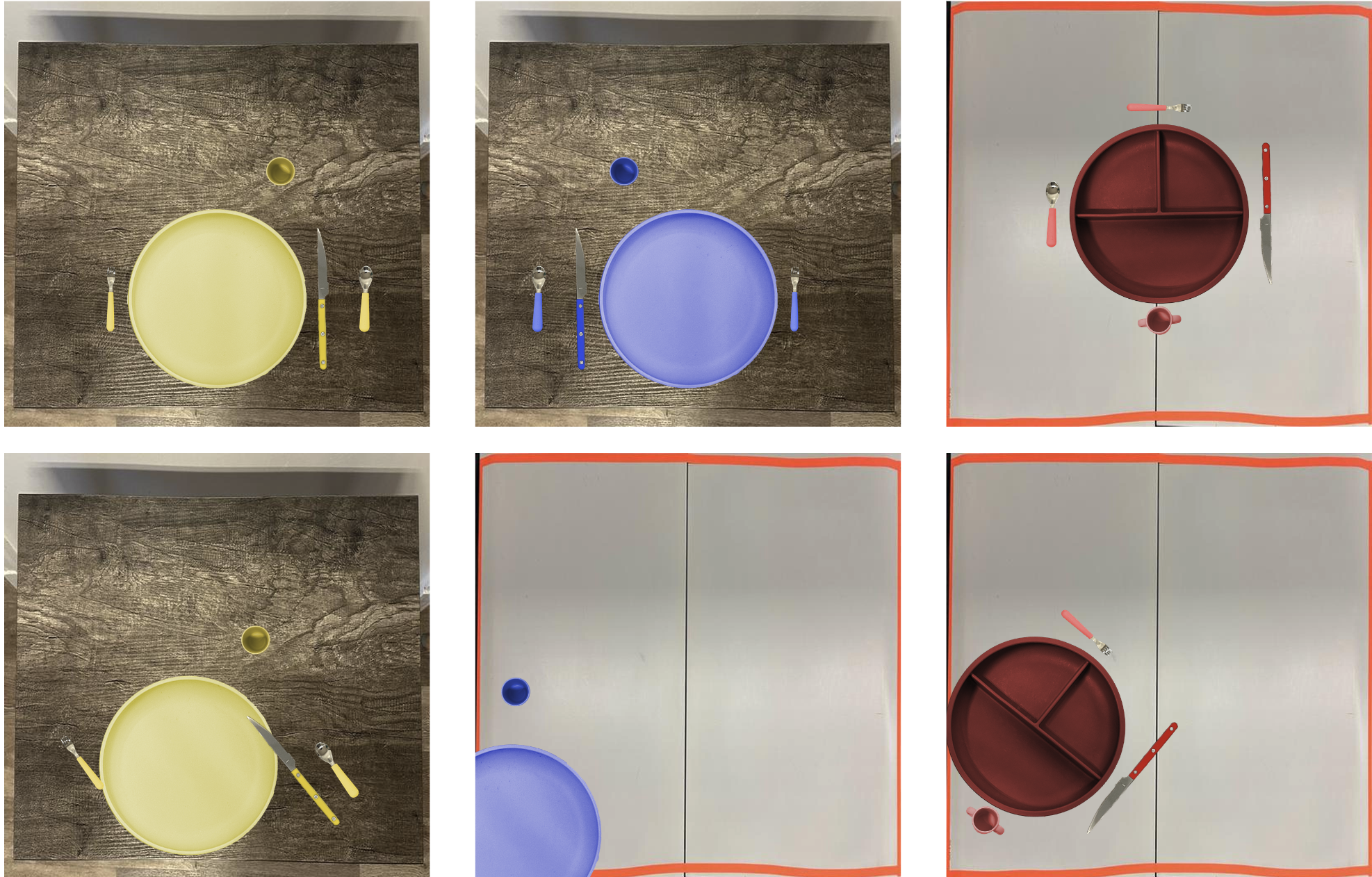}
    \caption{Examples of the simulated preferences over a variety of colors prior to being transformed (top) and after being transformed, bottom.}
    \label{fig:sim_examples}
\end{figure*}

Figure~\ref{fig:sim_examples} shows three examples of our simulated preferences before (top) and after (bottom) their transformations. We simulate a traditional Western table arrangement (center), a mirrored arrangement (center), and an abstract arrangement (right) to capture a variety of different difficulties. We add further difficulty by adding random transformations and rotations to the arrangements (bottom). 

\section{Data Collection Interface}

\begin{figure*}[t]
    \centering
    \includegraphics[width=0.75\linewidth]{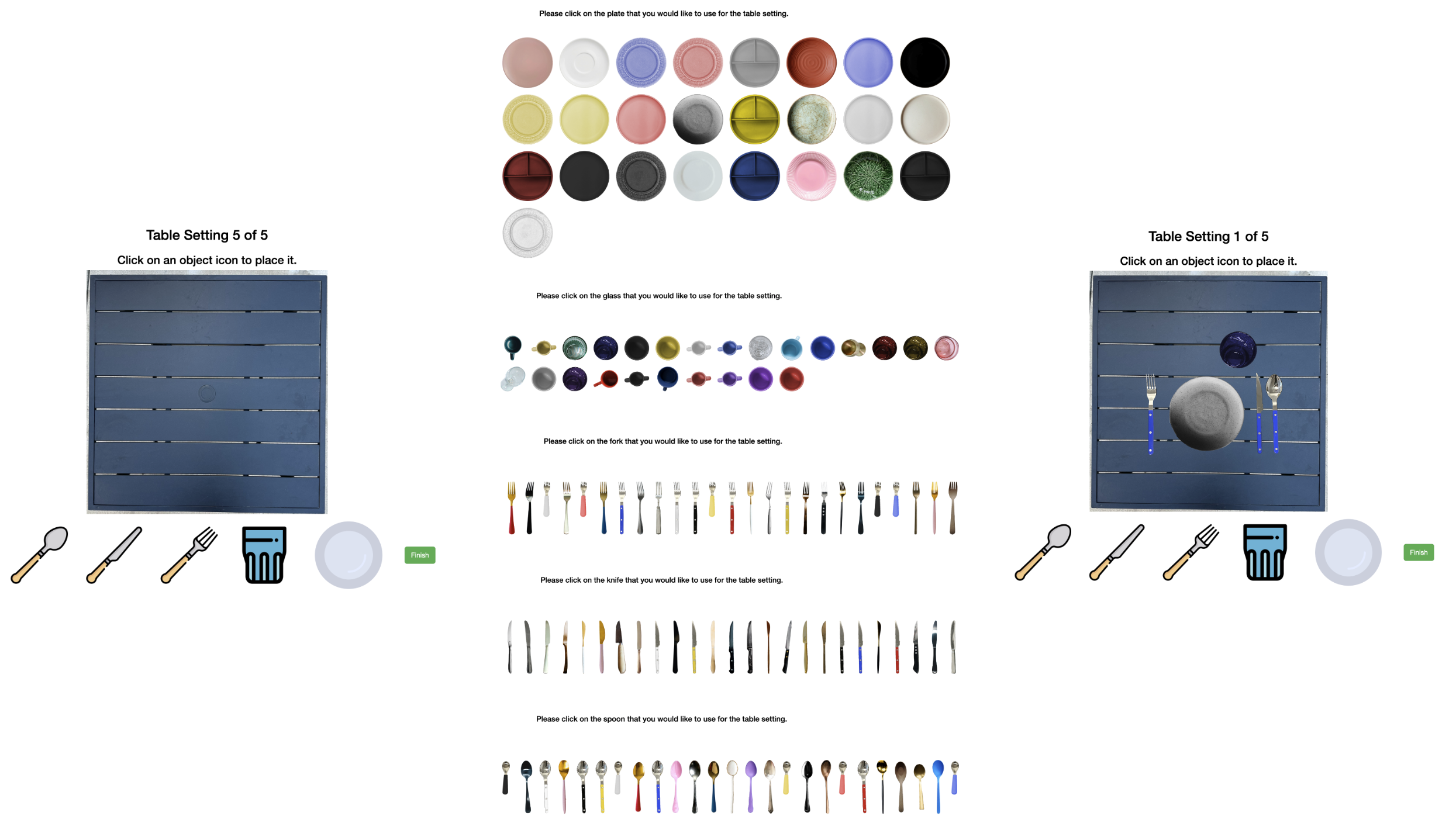}
    \caption{Images from the data collection interface. Participants are first presented with a n empty table and a set of object icons (left). After selecting an icon they are presented with the objects they can use to create the arrangement (center). After selecting an object, they can place it on the table. They continue this loop until they are satisfied with the arrangement (right). }
    \label{fig:interface}
\end{figure*}

Figure~\ref{fig:interface} shows images from the data collection interface. Participants are presented with a randomized order of empty tables they are asked to set. They must set a minimum of three objects. They first choose an object icon that represents the object they would like to place, then they are presented with all object instances of that category. After choosing an instance, they drag, drop and rotate the object on the table's surface. They continue until they are satisfied with the arrangement. We filter out participants who do not complete teh self evaluation portion of the task, or those who do not complete all 5 trials. 

\begin{figure*}[t]
    \centering
    \includegraphics[width=1\linewidth]{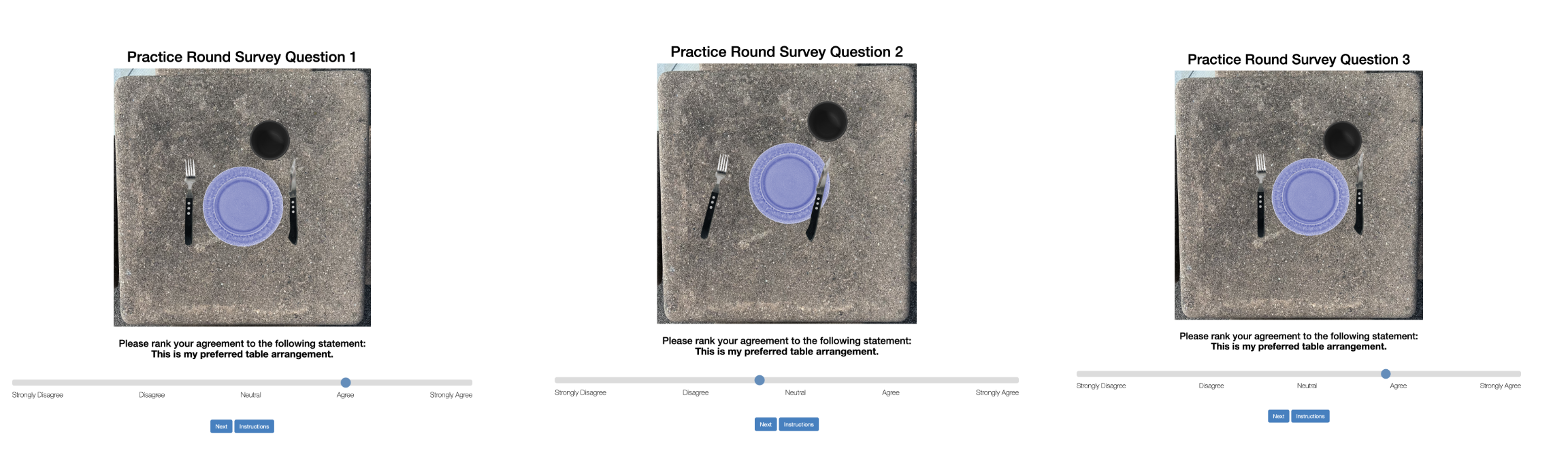}
    \caption{Images from the self evaluation portion of the user study. Participants rated their scene, left, then the scene was jittered and rated, center, finally the participants corrected the scene and rated again.}
    \label{fig:interface}
\end{figure*}

Figure~\ref{fig:interface} shows the self evaluation portion of the user study, where participants rated their scene and possible transformations of their scene that could represent positional errors our model could make. The goal of this section of the user study was to correlate user acceptability to random (jitters) and nonrandom (corrections) errors as measured in RMSD. This gives us an upper (i.e. if all model errors were nonrandom) and lower bound (i.e. if all model errors were random) against which to measure our models performance. We report the stricter (e.g. assuming all model errors are random) of these two measures in the paper. 

\textbf{Correcting the jittered scene.} Finally, we seek an objective measure of the types of errors that people find acceptable. To do this, we ask participants to correct the jittered scene back to its pre-jitter state. This creates an arrangement $a_{correc}$. Finally, we ask for an acceptability rating of $a_{correct}$, $b_{correct}$. 

To measure objective acceptability we perform a similar process by taking the difference in acceptability ratings between the post-correction arrangement and the initial arrangement: $b_{correct} - b_{initial}$ and correlating this value to the distance between the corrected scene and initial scenes: $d(a_{correct}, a_{initial})$. 

\subsection{Descriptive Self Evaluation Results}

\begin{figure*}[t]
    \centering
    \includegraphics[width=1\linewidth]{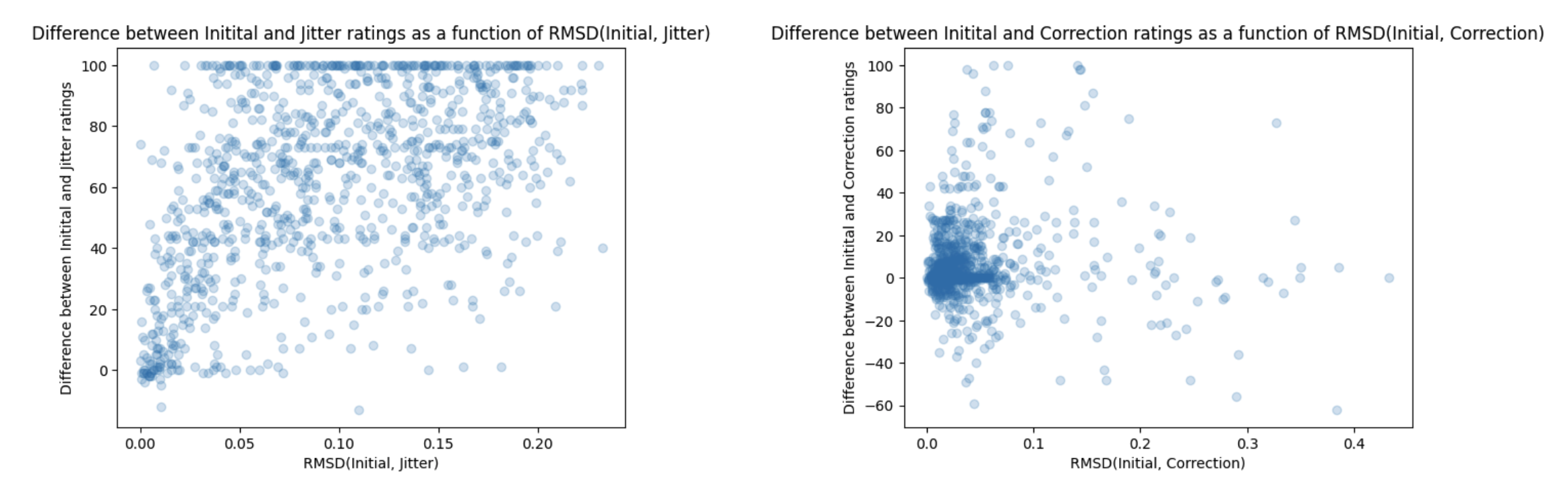}
    \caption{Plots of the difference between jitter rating and baseline rating (left) and correction rating and baseline rating (right). }
    \label{fig:descriptive}
\end{figure*}

Figure~\ref{fig:descriptive} shows the correlation between ratings differences and RMSD for both the jitter condition (representing random errors), left, and the correction condition (representing nonrandom errors), right. From the graph on the left, we see a cluster of participants close to the origin, and then a quick slope up, with many participants distributed througout the top of the graph. This shows that participants are very sensitive to random errors, which is the metric against with we report. We see from the graph on the right that participants typically correct the scene back to an accptable range, even with some error from the initial scene. 

\subsection{Example Arrangements}

\begin{figure*}[t]
    \centering
    \includegraphics[width=1\linewidth]{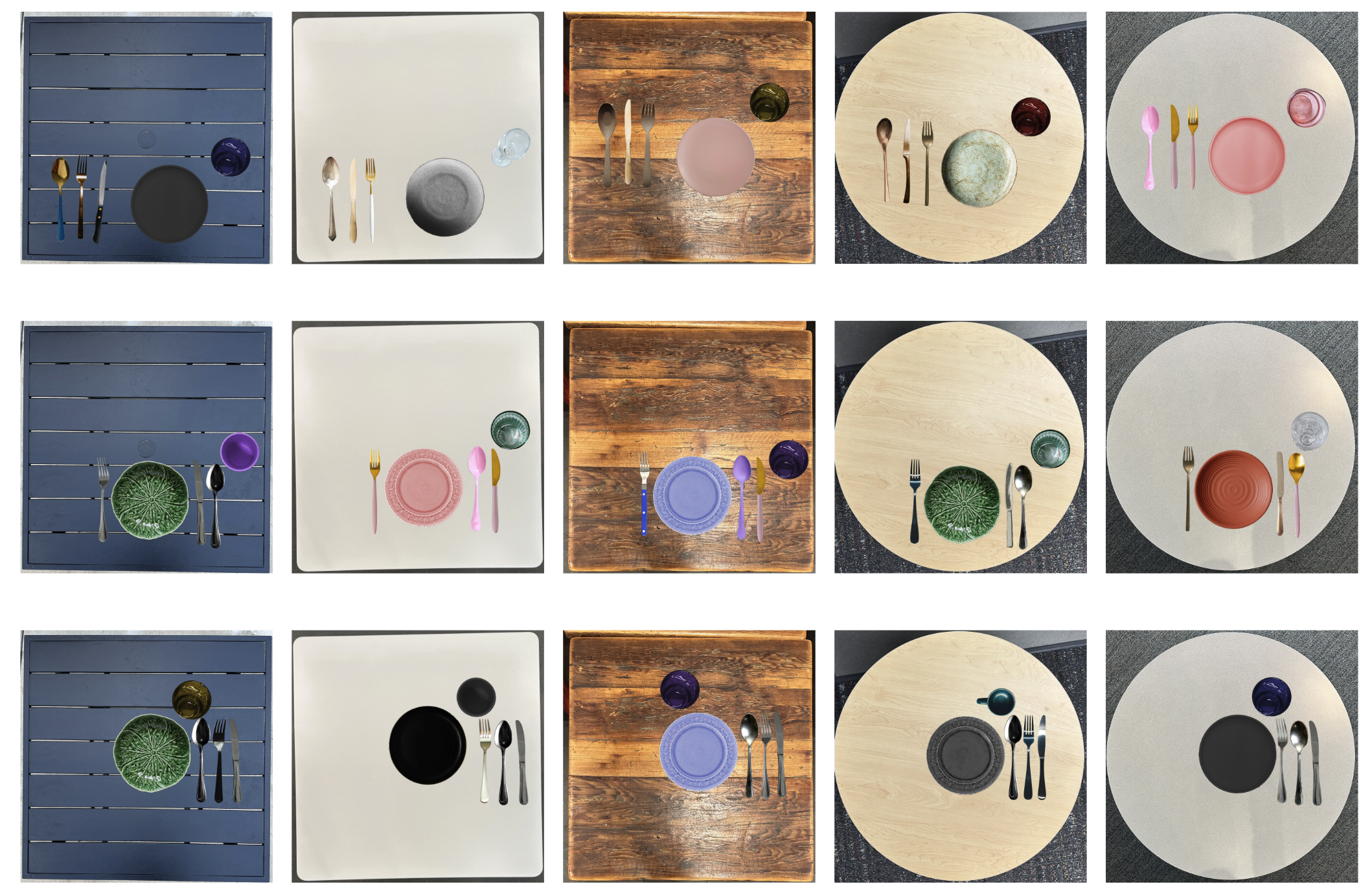}
    \caption{Example arrangements from three different participants. }
    \label{fig:examples}
\end{figure*}

Figure~\ref{fig:examples} shows five arrangements examples from three different participants. We can see a diversity across the arrangements with which objects they choose, where they place the objects on the table, and where they place the objects in relation to one another. We can also see some consistency within an arrangement across those same features. Our data set contains 995 of such arrangements drawn from 199 different people. 

\end{document}